\documentclass[conference]{IEEEtran}
\IEEEoverridecommandlockouts

\usepackage[pdftex]{graphicx}
\usepackage{epstopdf}
\usepackage[utf8]{inputenc} 
\usepackage[T1]{fontenc}    
\usepackage{booktabs}       
\usepackage{amsfonts}       
\usepackage{nicefrac}       
\usepackage{microtype}      
\newenvironment{proof}[1][Proof]{{\it #1. } }{\ \rule{0.5em}{0.5em}}
\usepackage{graphics, theorem, times, cite}
\usepackage{graphicx}
\usepackage{subfig}

\usepackage{tikz}
 
\usepackage[english]{babel}
\usepackage{ifpdf}
\usepackage{bm}

\ifpdf
	\usepackage[pdftex]{graphicx}
	\graphicspath{{./figures/}}
\else
	\usepackage[dvips]{graphicx}
	\graphicspath{./figures/}
\fi
\usepackage{color}
\usepackage{pgf, tikz, pgfplots, epstopdf}
\pgfplotsset{compat=1.18}
\usetikzlibrary{shapes, arrows, automata}
\usepackage{subfig}
\usepackage{amsmath}
\usepackage{amsfonts, amssymb,  bbm}
\usepackage{mathrsfs}
\usepackage{upgreek}
\usepackage{amsmath}
\DeclareMathOperator*{\argmax}{arg\,max}

\usepackage{algorithm,algpseudocode}
\usepackage{enumerate}
\usepackage{multirow}
\usepackage{rotating}
\usepackage{graphicx}    
\usepackage{caption}     
\usepackage{subcaption}  
\usepackage{url}

\usepackage{needspace}

\input{mySymbol.sty}

\usepackage{booktabs}       
\usepackage{nicefrac}       
\usepackage{microtype}      
\usepackage{xcolor}         

\newtheorem{theorem}{\hspace{0pt}\bf Theorem}

\newcommand{\QED}{\hfill\ensuremath{\blacksquare}}

\definecolor{mutedblue}{RGB}{70, 130, 180}   
\definecolor{mutedred}{RGB}{178, 34, 34}     
\definecolor{mygreen}{RGB}{34, 139, 34}  

\usepackage[  
    colorlinks = true,       
    pdfborder  = {0 0 0},    
    urlcolor   = mutedblue,  
    linkcolor  = mutedred,   
    citecolor  = mygreen]    
   {hyperref}

\title{Limit Analysis of Graph Neural Networks with Wireless Conflict Graphs
}

\author{Romina Garcia Camargo \quad Zhiyang Wang \quad Alejandro Ribeiro \thanks{RGC and AR are with the Department of Electrical and Systems Engineering, University of Pennsylvania, Philadelphia, PA (emails: \{rominag, aribeiro\}@seas.upenn.edu). ZW is with Halıcıoğlu Data Science Institute, UCSD, La Jolla, CA (email: zhw135@ucsd.edu).} }

\usepackage[dvipsnames]{xcolor}

\newcommand{\rofs}[1]{#1 \odot \big[\, \mathbf{1} - \bbA_m#1    \,\big]_+}

\def \rst {\rofs{\bbp(t)}}
\begin{document}
\pagestyle{plain}
\maketitle

\begin{abstract}
Graph Neural Networks (GNNs) have emerged as a powerful tool for wireless resource allocation that leverages the underlying graph structure of communication networks. Their transferability property enables models trained on small-scale graphs to generalize to large-scale deployments with little performance deterioration, a desirable property for currently growing networks. Wireless networks are sparse regimes, where a single node is connected to a small number of other users. This work establishes theoretical results for transferability of GNNs over graphs derived from sparse Random Geometric Graphs (RGGs). In particular, we focus on conflict graphs of RGGs used to model interference among links. Our approach considers the closeness between RGGs and Deterministic Grid Graphs (DGG) to establish bounds in the performance loss when a model is transferred across scales. We validate our theoretical findings through the problem of link scheduling, demonstrating that our learned policies consistently outperform existing benchmarks at scale. Finally, we examine the impact of our theoretical assumptions on empirical performance.
\end{abstract}


\section{Introduction}
\label{sec:intro}
Learning-based solutions are promising tools for addressing complex resource allocation challenges in next-generation networks. Graph Neural Networks (GNNs) are well-suited for these tasks, as the architecture exploits the underlying graph structure of wireless systems \cite{randall2025userassociation, wang2022learning, camargo2025wirelesslinkschedulingstateaugmented}. 
In particular, it allows for transferability:  models trained on small graphs generalize to larger networks with minimal performance loss.

While empirical studies show that GNNs successfully transfer in wireless settings \cite{eisen2020optimal}, these observations lack a formal foundation. Multiple works have explored the transferability of Graph Neural Networks \cite{ruiz2021transferability, MASKEY202348, keriven2020convergence, wang2024geometric, le2023sparsetransferability}. However, most existing frameworks typically rely on limit objects such as graphons \cite{ruiz2021transferability} or manifolds \cite{wang2024geometric}, which define transferability as connectivity increases toward a dense limit. Physical constraints in wireless channels limit the number of connections per user, resulting in sparse graphs that traditional dense-limit theories fail to account for. To bridge this gap, recent work has begun analyzing transferability on sparse graphs via graphops \cite{le2023sparsetransferability} or using Random Geometric Graphs (RGGs), leveraging the geometric structure intrinsic to communication systems \cite{camargo2025graphneuralnetworkslarge}.

Resource allocation tasks such as link scheduling depend on the relationships between communication links rather than the users themselves. This is often tackled via conflict graphs, where links are modeled as nodes and edges represent interference or shared resources \cite{linkschedulingusinggnns, cao2020resourceallocationultradense} (Section \ref{sec:problemform}). However useful, it creates a specific sparse regime distinct from the underlying physical network. 

In this work we establish the theoretical analysis for transferability of GNNs on conflict graphs (Sections \ref{sec:learningbytransf} and \ref{sec:transf}). Extending our framework \cite{camargo2025graphneuralnetworkslarge}, we model wireless connectivity using Random Geometric Graphs (RGGs) and prove that transferability holds via Deterministic Grid Graphs (DGGs). By leveraging their spatial periodicity, we overcome the analytical challenges posed by the topological irregularities of random node placements. Our results demonstrate that as the number of nodes increases, transferability is guaranteed for conflict graphs of RGGs that approximate conflict graphs of DGGs.

We validate our theoretical framework with numerical experiments on the wireless link scheduling problem, incorporating a minimum transmission requirement per link (Section \ref{sec:simu}). The GNN is trained on conflict graphs derived from RGGs to select which links should be scheduled at each time step. Empirical results demonstrate that our transferred model consistently outperforms benchmarks at larger scales. Furthermore, we characterize the robustness of our RGG-DGG proximity assumption by analyzing model performance as the graph topology deviates from a grid-like structure.

\section{Resource Allocation on Conflict Graphs}
\label{sec:problemform}
\begin{figure*}
    \centering
 \includegraphics[width=0.95\linewidth]{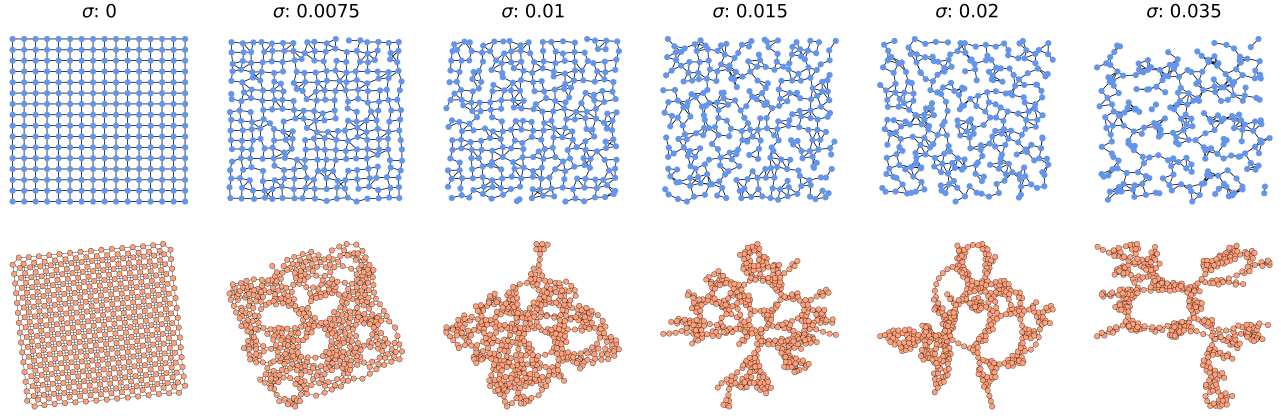}
   \caption{Illustration of Random Geometric Graphs (blue) and their corresponding conflict graphs (pink) as the noise in positions increases. Observe that the RGG deviates from a grid-like structure as the positional noise increases.}
    \label{fig:large_graph_stability_illustration}
\end{figure*}

Let $\bbG=(\ccalV, \ccalE)$ be a graph with $|\ccalV|=v$ nodes and $|\ccalE|=e$ edges. Its conflict graph $\bbG_C=(\ccalV_C, \ccalE_C)$ will have $|\ccalV_C|=e$ nodes, one for each edge in $\bbG$. An edge $(i, j)\in\ccalE_C$ will exist if edges $i$ and $j$ are incident on the same node in $\bbG$. Figure \ref{fig:large_graph_stability_illustration} shows an illustration of graphs and their conflict graphs. 

In wireless communication settings, conflict graphs are widely used to model resource allocation tasks (Section \ref{sec:optimalresall}, \cite{eisen2020optimal}), as they represent dependencies or forbidden combinations between the edges of the original graphs \cite{linkschedulingusinggnns, cao2020resourceallocationultradense}.  In particular, these models are used to analyze the links of the network, such as via the primary interference model for wireless link scheduling \cite{hajekb88} (Section \ref{sec:wls}).

\subsection{Optimal Resource Allocation in Wireless Settings}
\label{sec:optimalresall}
Let there be a network with $m$ users, where each user is associated with a state at time $t$ given by the vector $\mathbf{x}_m(t)\in\mathbb{R}^m$. Given the channel gains in $\mathbf{S}_m(t)\in\mathbb{R}^{m\times m}$, where $[\bbS_m(t)]_{ij}$ is the gain from user $i$ to user $j$, we are interested in allocating resources across the network. Let $\mathbf{p}(\bbx_m(t),\bbS_m(t))$ be a mapping from network states to the instantaneous resource allocation policy across the network. The system will then return a reward given the allocation, $\mathbf{f}(\mathbf{p}(\bbx_m(t),\bbS_m(t));\mathbf{x}_m(t),\mathbf{S}_m(t))$. Considering the expectation over the stationary joint distribution of $(\mathbf{x}_m,\mathbf{S}_m)$, we focus on the long-term average performance as
\begin{align}
    \mathbf{r}_m = \mathbb{E}[\mathbf{f}
    (\mathbf{p}; \mathbf{x}_m, \mathbf{S}_m)]. \label{eq:reward}
\end{align}

Consider a utility function $u_0(\mathbf{r}_m)$ and long-term system constraints $\mathbf{u}(\cdot)$. Our goal is to maximize the expected reward \eqref{eq:reward} by designing our allocation policy $\bbp(\mathbf{x}_m,\mathbf{S}_m)$:
\begin{align}
 \mathbf{p}^\star(\mathbf{x}_m, \mathbf{S}_m) = & \argmax_{\mathbf{p}(\mathbf{x}_m,\mathbf{S}_m) \in \mathcal{P}(\mathbf{x}_m,\mathbf{S}_m)} \ 
    u_0(\mathbf{r}_m), \label{eq:pa}
    \\
    \text{s.t.} \ &\mathbf{r}_m =\mathbb{E}[\mathbf{f}(\mathbf{p}(\mathbf{x}_m,\mathbf{S}_m); \mathbf{x}_m,\mathbf{S}_m)], \nonumber\\
  &\mathbf{u}(\mathbf{r}_m) \geq\mathbf{0}. \nonumber
\end{align}
The objective in Equation \eqref{eq:pa} is often nonconvex in $\mathbf{p}$, which makes finding a solution challenging. The common approach introduces a parameterized policy $\bm{\Phi}(\mathbf{x}_m,\mathbf{S}_m; \mathbf{H}_m)$,
with parameters $\mathbf{H}_m\in\mathbb{R}^{s}$ and learns a solution (Section \ref{sec:learningbytransf}). 

\subsection{Example: Wireless Link Scheduling}
\label{sec:wls}
Let $\bbG$ be wireless network with $m$ links, where $p_i(t)\in\{0,1\}$ denotes the status of link $i\in\{1,2,\dots, m\}$.  Entry $p_i(t) = 1$ indicates that the link $i$ is scheduled to transmit at time step $t$. Two links interfere each other if they are scheduled at the same time and share a common user \cite{hajekb88}. The vector $\bbp(t) = [p_i(t)]_{1\leq i\leq m}\in \{0,1\}^m$ describes the scheduling status of all links and is therefore supported on the edges of $\bbG$. 

Let us define $\bbG_C$ the conflict graph of $\bbG$, then $\bbp(t)$ is supported on the nodes of $\bbG_C$. We denote the adjacency matrix of $\bbG_C$ as $\bbA_m\in\{0,1\}^{m\times m}$. 
The successful transmissions $\bbr_m(\bbp)$ resulting from a schedule $\bbp(t)$ can be computed via the Hadamard product between the schedules and nonnegative projection $[\mathbf{1}  - \bbA_m \bbp(t)]_+$,
\begin{align}\label{eqn_rate_pointwise}
    \bbr_m(\bbp) = \rst.
\end{align}

We are interested in the average rate over a time horizon. Wireless link scheduling seeks a policy that maximizes the long-term average of the sum rate $\bbr_m(\bbp)$ over $T$ time steps. The policy must also satisfy the interference constraints intrinsic to the topology of the network. 
We incorporate a minimum transmission requirement for each link, $\bm\Delta \in \reals_+^{m}$, such that all links must transmit at least a fraction of time:
 \begin{alignat}{3} \label{eq:wls}
     \bbp^\star(1: T) 
         = & \argmax_{ \bbs(t)\in\{0,1\}^K}  
                   && \frac{1}{T} \sum_{t=1}^{T}  \bbone^\top \rst  , 
                            \nonumber \\
             & \text{s.t.}  
                   && \frac{1}{T} \sum_{t=1}^{T} \rst
                            \geq \bbDelta.
 \end{alignat}

To solve this constrained optimization task we turn to the Lagrangian dual domain, defining a new learning objective that combines our objective function and a penalty term for constraint violations. Nonetheless, Lagrangian maximizers will not result in workable policies as they will not vary over time. By leveraging the dual descent dynamics, we can incorporate the evolution of the dual variable $\bblam$ into our learning via state augmentation \cite{calvofullana2023stateaugmentedconstrainedreinforcement}. More details about the problem formulation and the algorithm can be found in \cite{camargo2025wirelesslinkschedulingstateaugmented}.

\section{Training by Transference}
\label{sec:learningbytransf}
Finding optimal solutions to resource allocation problems such as the one in Equation \eqref{eq:pa} is often computationally prohibitive. Common approaches approximate these solutions by using a parameterized policy $\bm{\Phi}(\mathbf{x}_m,\mathbf{A}_m; \mathbf{H}_m)$. The objective is to identify the optimal parameters $\mathbf{H}^\star_m\in\mathbb{R}^{s}$ that maximize the network utility  while satisfying the expected constraints: 
\begin{align}\label{eq:parameterized}
 \mathbf{H}_m^\star = & \argmax_{\mathbf{H}_m\in\mathbb{R}^{s}} \ 
    u_0(\bbr_m), 
    \\
    \text{s.t.} \ &\mathbf{r}_m = \mathbb{E}[\mathbf{f}(\bm{\Phi}(\mathbf{x}_m,\mathbf{A}_m; \mathbf{H}_m); \mathbf{x}_m,\mathbf{A}_m)], \nonumber\\
  &\mathbf{u}(\mathbf{r}_m)  \geq\mathbf{0}. \nonumber
\end{align}

In wireless applications, Graph Neural Networks (GNNs) are the preferred architecture for $\bbPhi$. By training over diverse channel realizations, GNNs learn efficient allocation policies that maintain awareness of the network topology.

Let the graphs used for training have $m$ nodes, and assume we want to execute on graphs with $n$ nodes, such that $m <n$. In this case, the evaluation consists of finding policies $\bbPhi(\bbx_n, \bbA_n; \bbH_m^\star)$. Note that while the descriptions of the network state ($\bbx_n, \bbA_n$) now have $n$ dimensions, the parameter matrix $\bbH_m^\star$ is independent of the number of nodes in the graph. This implies we \textit{can} execute a GNN trained on a small graph on a larger graph, i.e. \textit{transfer} the GNN. Nonetheless, it is necessary to assess whether these policies remain optimal as the network scales. To analyze this, we first present the GNN architecture in Section \ref{sec:gnnss}. Furthermore, we introduce Random Geometric Graphs and their deterministic grid counterparts (Section \ref{sec:rgg}) to later present our transferability results (Section \ref{sec:transf}).

\subsection{Graph Neural Networks}
\label{sec:gnnss}
A Graph Neural Network is composed of a cascade of layers, each containing a graph convolutional filter and a pointwise nonlinearity $\gamma$, with $\gamma : \mathbb{R} \rightarrow \mathbb{R}$. Given a graph signal $\mathbf{x}\in\mathbb{R}^m$, a graph convolutional filter is defined as follows: 
\begin{align}
\label{eq:graphfilter}
\mathbf{y} = \sum_{k=0}^{K-1} h_{k} \mathbf{A}^k \mathbf{x},
\end{align}
where $K$ is the order of the filter and $\{h_k\}_{k=0}^{K-1}$ are the coefficients. The filter is a polynomial on a matrix representation of the graph, $\mathbf{A}\in\mathbb{R}^{m\times m}$. 

A single-layer GNN applies the nonlinearity $\gamma$ to the output of the graph filter in Equation \eqref{eq:graphfilter}. For a GNN with $L$ layers, the $l$-th layer takes $\mathbf{x}_{l-1}$, the output of the previous layer, as input signal. This is later processed via $\gamma$ to obtain $\mathbf{x}_{l}$:
\begin{align}
\mathbf{x}_l = \gamma\left(\sum_{k=0}^{K-1} h_{lk} \mathbf{A}^k \mathbf{x}_{l-1}\right).
\end{align}

We can summarize the weights in a matrix $\mathbf{H} \in \mathcal{H}$ as the trainable parameters of the network. 

\subsection{Deterministic and Random Geometric Graphs}
\label{sec:rgg}
Define a metric space of size $W\times W$ and place $v$ nodes randomly following a uniform distribution in the space, $\mathcal{V} \sim\mathcal{U}^2(0, W)$. We create an edge between any two nodes that meet that their Euclidean distance is at most a fixed radius $r$, $\mathcal{E} = \{(i,j) : d(i,j) \leq r\}$. These sets define a Random Geometric Graph (RGG) \cite{penrose2003random}, $ \bbG = (\mathcal{V}, \mathcal{E})$, which appear naturally in wireless settings due to the relation between the fixed radius $r$ and connectivity determined by signal strength. 

Define a regular lattice in a Euclidean space with $v$ intersections. By placing a node in each intersection and creating links between neighboring nodes following the grid, we can obtain a deterministic grid graph (DGG), $\tdG = (\tilde{\mathcal{V}}, \tilde{\mathcal{E})}$. 

RGGs can be seen as a perturbation of DGGs, resulting from adding Gaussian noise $\eta \sim \mathcal{N}(0,\sigma^2)$ to the node positions. If the perturbation is small, an analysis can be done that seamlessly relates RGGs and DGGs \cite{camargo2025graphneuralnetworkslarge}. As observed in Figure \ref{fig:large_graph_stability_illustration} (see $\sigma=0$), the conflict graph of a DGG will also be highly regular and can be easily related to the conflict graph of a RGG. This forms the basis of the theoretical analysis we present in Section \ref{sec:transf}. 

\section{Transferability in Conflict Random Geometric Graphs}
\label{sec:transf}
Consider a DGG $\tdG$ with $m_1$ edges. Construct a RGG $\bbG$ by adding positional noise to $\tdG$, connecting the nodes given a fixed radius such that it has $m_2$ edges with $m_1\simeq m_2$. In order to study the transferability of GNNs from conflict Random Geometric Graphs to conflict Deterministic Grid Graphs, we consider the normalized adjacency matrices of conflict graphs ${\tilde{\bbA}}_{m_1}\in\mathbb{R}^{m_1\times m_1}, \bbA_{m_2}\in\mathbb{R}^{m_2\times m_2}$. 

To assess the similarity between a DGG and a RGG, we are interested in comparing their adjacency matrices. To enable the comparison we zero-pad the normalized adjacency matrices to obtain $\bbA, \tilde{\bbA}\in\mathbb{R}^{(m_1+m_2)\times (m_1+m_2)}$. By embedding both graphs into a spatially-aligned vector space, the perturbation captures only the mismatched edges between the two topologies.

We assume that GNN consists of layers cascading the graph filter, which is assumed to be integral Lipschitz continuous and the normalized Lipschitz point-wise nonlinearities \cite{ruiz2021gnnarch}.
We note that these are reasonable assumptions and have been implemented widely in theory and in practice.

With these definitions, we present in Theorem \ref{the:rgg-gg-gnntransf} the result for transferability of GNNs between RGG and DGG.
\begin{theorem}\label{the:rgg-gg-gnntransf}
    Let $\tdG$ be a DGG with $m_1$ edges and $\bbG$ be a RGG with $m_2$ edges. Define $\bbA, \tilde{\bbA} \in \{0,1\}^{(m_1+m_2)\times (m_1+m_2)}$ the zero padded adjacency matrices of their conflict graphs, such that $\|\bbA-\tilde{\bbA}\|\leq \varepsilon$. Let $\bbPhi(\mathbf{x},\mathbf{A};\mathbf{H})$ be a $L$-layer GNN with a graph filter $\bbH$ that is integral Lipschitz with constant $C$ and a normalized Lipschitz nonlinearity $\gamma$. The GNN can be transferred between $\tilde{\bbA}$ and $\bbA$ with minimal performance loss:
\begin{align}
    \|\bbPhi(\bbx, \tilde{\bbA}; \bbH)-\bbPhi(\bbx, \bbA; \bbH)\| \leq& 2L\sqrt{\varepsilon C}\|\bbx\|.
\end{align}
\end{theorem}
\begin{proof}
    See Appendix \ref{app:the1}.
\end{proof}

Theorem \ref{the:rgg-gg-gnntransf} shows that the error between running a GNN on a DGG and a RGG is small, provided the graphs are close. 

We then analyze the transferability of GNN trained on a DGG of size $m$ to a DGG of size $n$, with $m<n$. This is an extension of the transferability studies of CNNs \cite{owerko2023transferabilityconvolutionalneuralnetworks}. Let these two graphs be described via their normalized adjacency matrices ${\tilde{\bbA}}_m\in\mathbb{R}^{m\times m}, {\tilde{\bbA}}_n\in\mathbb{R}^{n\times n}$. 

Let $\ccalL({\tilde{\bbA}}_m,\bbH)$ be the performance measure associated with running a GNN on ${\tilde{\bbA}}_m$ \cite{owerko2023transferabilityconvolutionalneuralnetworks}. We assume an input $\bbx$ and compare the GNN to the optimal policy $\bbp^*$. 
\begin{align}
    \ccalL({\tilde{\bbA}}_m,\bbH) = \frac{1}{m}\|\bbPhi(\bbx, {\tilde{\bbA}}_m; \bbH)-\bbp^*\|^2
\end{align}
Similarly, let $\ccalL(\tilde{\bbA}_n,\bbH)$ be the performance when applied to ${\tilde{\bbA}}_n$. We compare the difference between the loss functions by considering a window of the large graph, ${\tilde{\bbA}}_n$, and comparing it to the small graph, ${\tilde{\bbA}}_m$.
\begin{theorem}\label{th:dggtondgg}
Let $\bbPhi(\bbx, {\tilde{\bbA}}_n; \bbH)$ be the parameterized policy that achieves a performance loss $ \ccalL({\tilde{\bbA}}_n,\bbH) $ when applied on a grid graph with size $n$ and achieves a loss of $ \ccalL({\tilde{\bbA}}_m,\bbH) $ when applied on another grid graph with size $m$. Suppose $n > m$, the difference of these two losses can be bounded as 
\begin{align}
    \ccalL({\tilde{\bbA}}_n,\bbH) \leq \ccalL({\tilde{\bbA}}_m,\bbH) + C\mathbb{E}[\bbx^2]+ 2\sqrt{\ccalL({\tilde{\bbA}}_m,\bbH)C\mathbb{E}[\bbx^2]},
\end{align}
where $C=\frac{H_K^2}{m}(2\sqrt{m}K+K^2)$ and $H_K=\sum_{k=0}^{K-1}|h_k|\|\tilde{\bbA}_m\|_2^k$.
\end{theorem}
\begin{proof}
    See Appendix \ref{app:the2}.
\end{proof}

In Theorem \ref{th:dggtondgg}, the performance difference between the evaluation on small and large DGGs is bounded by a small constant under mild assumptions. As $n$ increases, the interior nodes of both graphs have the same neighborhoods and the boundary effects go to zero.

A GNN trained on small RGGs can be transferred to small DGGs (Theorem \ref{the:rgg-gg-gnntransf}). Moreover, a GNN that performs close to optimality on small DGGs can maintain this performance for large DGGs (Theorem \ref{th:dggtondgg}). Finally, because performance in large DGGs will be similar to performance in large RGGs, it follows that a GNN can be trained in small RGGs and transferred to large RGGs with minimal performance loss as long as the RGGs can be seen as a shift of its corresponding DGGs with $\|\bbA-\tilde{\bbA}\|\leq \varepsilon$. We summarize this result in the following theorem:
\begin{theorem}\label{the:rggtonrgg}
    Under assumptions analogous to Theorem \ref{the:rgg-gg-gnntransf}, consider an L-layer GNN $\bbPhi(\mathbf{x},\mathbf{A};\mathbf{H})$ trained to minimize the loss $\ccalL(\bbA_m,\bbH)$ for RGGs of size $m$, such that $\ccalL(\bbA_m,\bbH)\leq\zeta$ and suppose the deviation of RGG to its corresponding DGG is bounded by $\epsilon$. The GNN can be transferred to graphs of size $n$ with minimal performance deterioration:
\begin{align}
&\nonumber | \ccalL({ {\bbA}}_n,\bbH)  - \ccalL({ {\bbA}}_m,\bbH) |= \\ 
    & \mathcal{O}\Bigg(\sqrt{\zeta}\left(\sqrt{\epsilon}\|\mathbf{x}_{n}\| + \sqrt{\epsilon}\|\mathbf{x}_{m}\|\right) + \epsilon\|\mathbf{x}_{n}\|^2  + \epsilon\|\mathbf{x}_{m}\|^2 \Bigg) 
\end{align}
\end{theorem}
\begin{proof}
   See Appendix \ref{app:the3}.
\end{proof}

\section{Numerical Experiments}
\label{sec:simu}
We conduct simulations on the wireless link scheduling problem as per the formulation in Equation \eqref{eq:wls}.\footnote{The code used to run the experiments and the implementation details are available for reproducibility \url{https://github.com/romm32/rgg_transferability}}. We observe the transferability of Graph Neural Networks over conflict graphs in practice, evaluating the scalability of the learned policies via comparisons with baselines at scale (Section \ref{sec:performance}). Furthermore, we analyze how robust the policies are to different levels of perturbation between a DGG and a RGG (Section \ref{sec:robustness}). 

We consider a GNN with $L=3$ layers, with leaky ReLU pointwise nonlinearities. The output signal of the last layer is passed through a sigmoid nonlinearity in order to obtain entries between 0 and 1. The performance of the algorithm is measured by binarizing the continuous output of the GNN with a threshold of 0.5 to decide which links to turn on. 

We consider $T=200$ time steps. The percentage of links transmitting a fraction lower than $\bbDelta=0.1$ of the time steps (i.e. violating the constraint), along with the long-term average percentage of links transmitting without interference (i.e. the objective function) are the two main performance metrics. All evaluations presented are on datasets unseen during training. 

\subsection{Performance at Scale}
\label{sec:performance}
\begin{figure}
    \centering
    \includegraphics[width=0.99\linewidth]{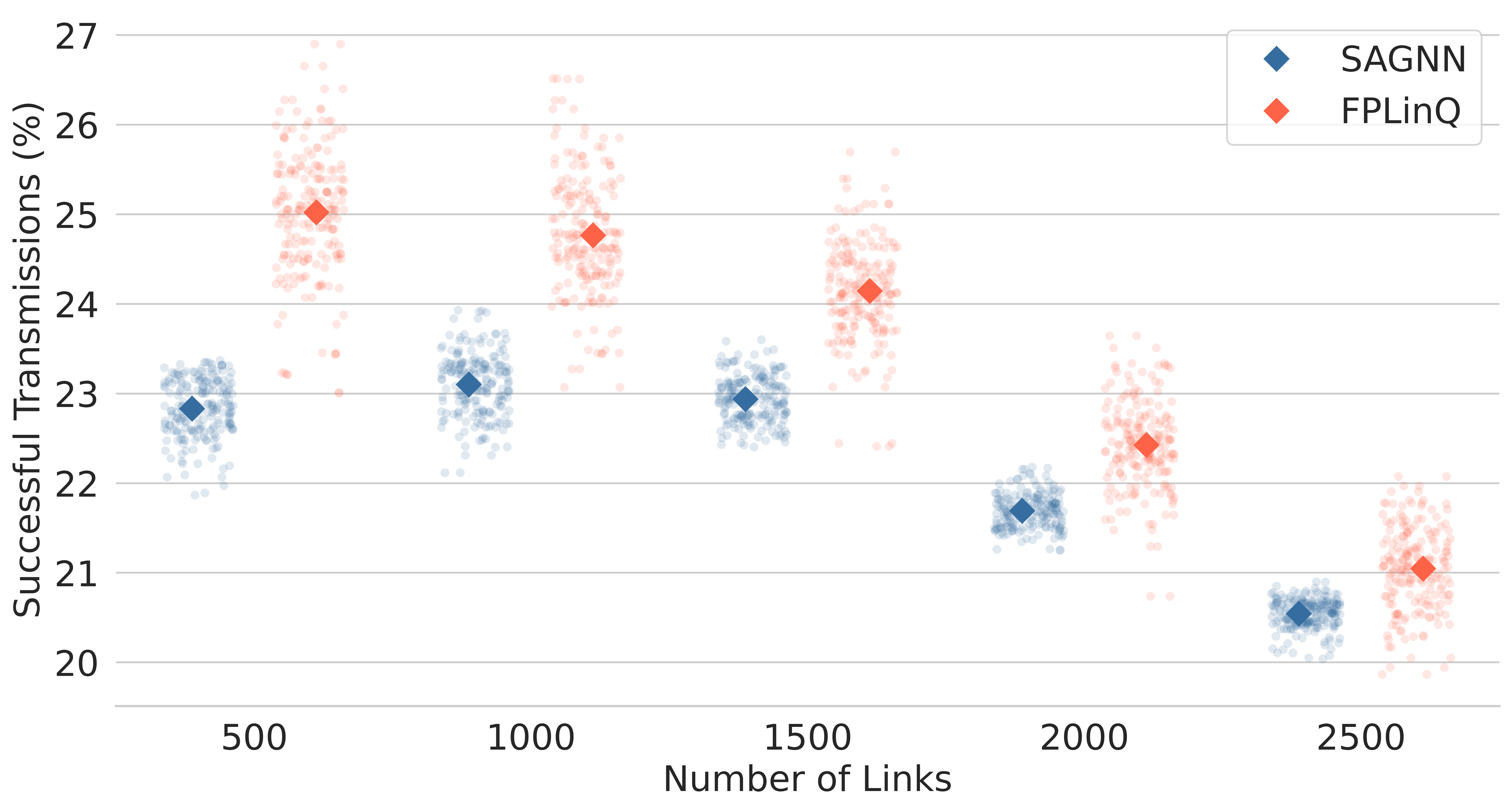}
    \caption{Comparison between the percentage of successfully scheduled links between SAGNN (ours) and FPLinQ. Our model is trained for graphs with $K\simeq500$ links and transferred to larger scales without significant performance loss.}
    \label{fig:scalabilityOF}
\end{figure}

The policy was trained using $100$ graphs with $K\simeq500$. The graphs are RGGs obtained via DGGs, adding positional noise with $\sigma=0.01$ (see Figure \ref{fig:large_graph_stability_illustration}). Figure \ref{fig:scalabilityOF} shows how our State Augmented GNN (SAGNN) transfers across scales. Performance does not degrade as we evaluate on five datasets with different scales, remaining close to 20-25\% of active links. Note that we present the average over 100 graphs on each dataset. The values achieved for the total average rates are close to the size of the maximum independent sets of the graphs, which represents the optimal solution to this problem. 

We compare our results with FPLinQ \cite{shen2017fplinq}, a fractional programming approach to solve link scheduling that focuses on instantaneous rate maximization. We perform close to this state-of-the-art baseline and achieve runtimes that are orders of magnitude more efficient (a speed up of 4x for $K\simeq500$ and an improvement of 30x for $K\simeq2500$). Furthermore, our GNN-based policy can be implemented in a distributed manner, allowing practical use.

\begin{figure}
    \centering
    \includegraphics[width=0.95\linewidth]{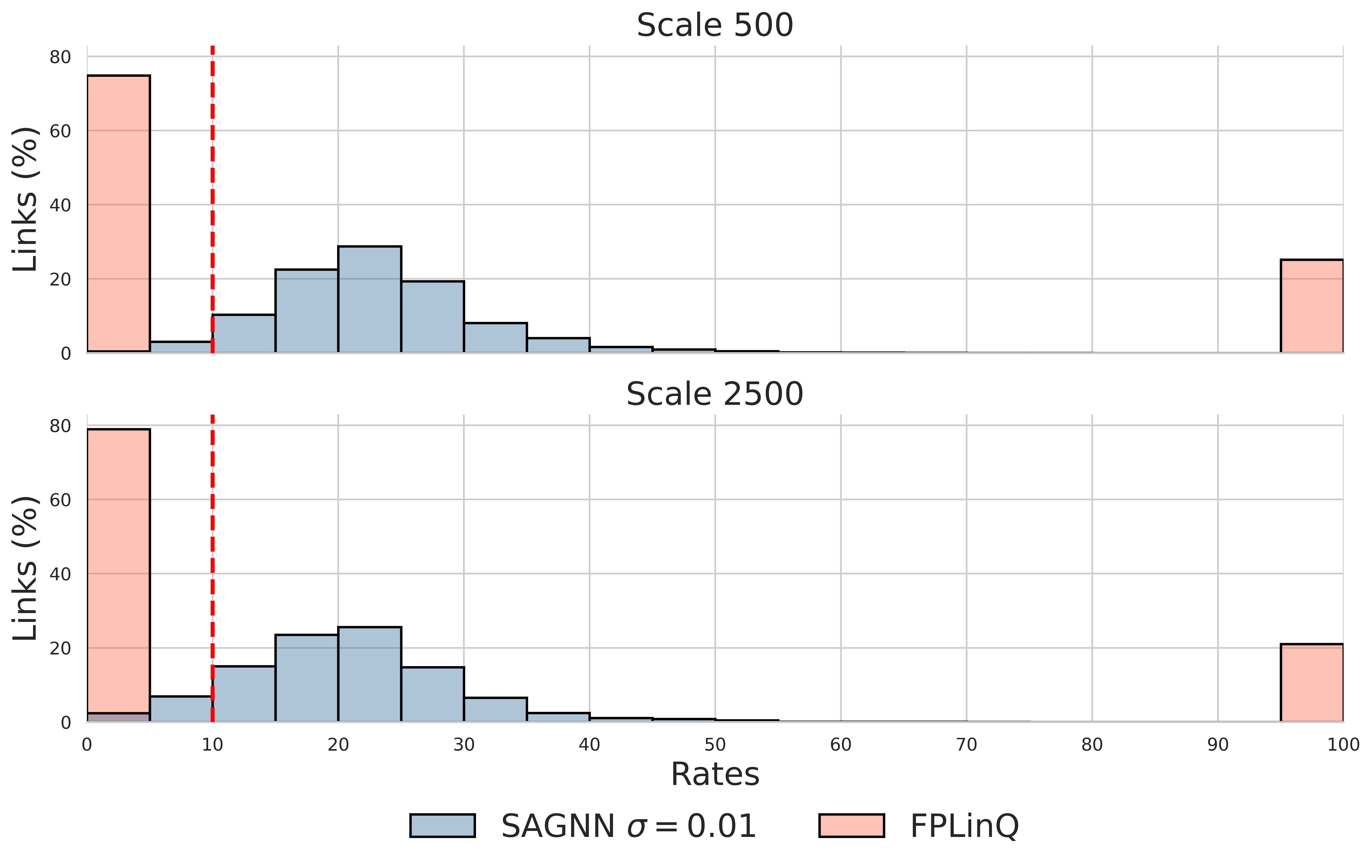}
    \caption{Average rates achieved by different links, presented as the percentage of time $T$ each link is scheduled. The minimum transmission requirement is set to $\bbDelta=10\%$ (red).}
    \label{fig:achievedrates}
\end{figure}

In Figure \ref{fig:achievedrates} we show a histogram of the average rates achieved by the different links, comparing both algorithms for scales $K=\{500, 2500\}$. SAGNN diversifies the schedules across all links, with the vast majority achieving rates higher than the $10\%$ transmission requirement. Nonetheless, the FPLinQ baseline repeatedly schedules the same links, not allowing for fair sharing of the resources.

\subsection{Robustness to Different Perturbation Levels}
\label{sec:robustness}
Our theoretical results in Theorem \ref{the:rggtonrgg} assume that the RGG must be close enough to a DGG, i.e. the perturbation level must be small. This can be translated in practice to adding positional noise with low variance $\sigma$. 

\begin{figure}
    \centering
    \includegraphics[width=1.1\linewidth]{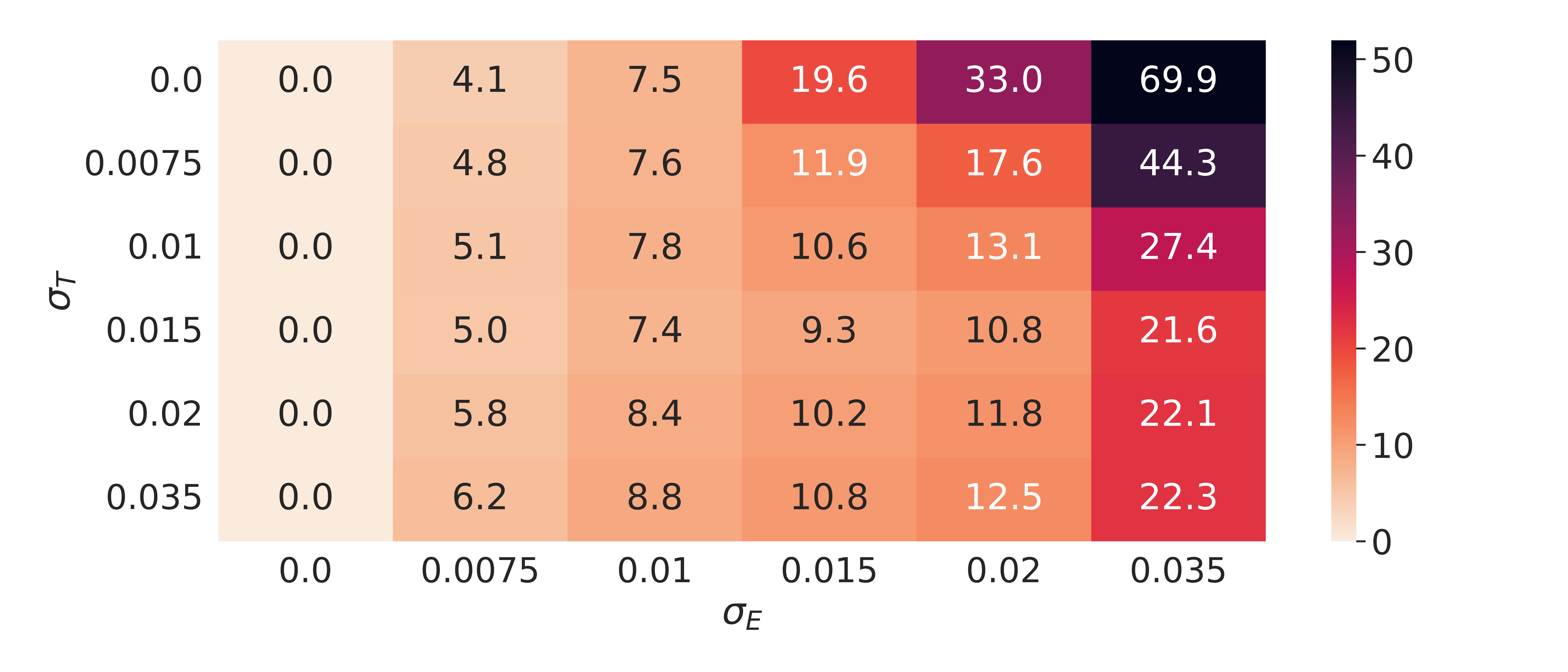}
    \caption{Study of the robustness of policies trained for a given level of position noise $\sigma_T$ and evaluated on $\sigma_E$. We present the percentage of links violating the constraint.}
    \label{fig:robustness}
\end{figure}

In Figure \ref{fig:robustness} we show the constraint violation results of training six models, one for each of the noise levels illustrated in Figure \ref{fig:large_graph_stability_illustration}. Each of these models is evaluated on six datasets containing 100 graphs each, generated by adding different levels of positional noise to a DGG. As expected, we see that models trained with lower noise levels ($\sigma_T=\{0, 0.0075, 0.01\}$) generalize worse to higher levels noise levels ($\sigma_E=\{0.015, 0.02, 0.035\}$). Models trained for larger noise levels achieve good performance across all noise levels. 

\section{Conclusions}
\label{sec:conclusion}
Random Geometric Graphs are a simple, yet accurate model of wireless networks. This work provides the theoretical study of transferability of Graph Neural Networks on conflict graphs derived from sparse RGG. This ensures wireless policies trained to satisfy requirements on the links of a network can be transferred to larger graphs with minimal loss in performance. We verified our theoretical results hold with numerical experiments on the wireless link scheduling task. It remains crucial to better assess the relevance of our assumptions with respect to practical development, which we leave for future work.

\urlstyle{same}
\bibliographystyle{IEEEtran}
\bibliography{references}

\section{Appendix}
\label{sec:appendix}
We provide the proofs for the different theoretical results in the work.

\subsection{Proof of Theorem 1}
\label{app:the1}
Before showing the transferability of GNNs from DGG to RGG, we present in Theorem \ref{th:rggtodggfilter} the transferability of graph filters from DGG to RGG.

\begin{theorem}\label{th:rggtodggfilter}
    Let $\tdG$ be a deterministic grid graph with $M$ edges and $\bbG$ be a random geometric graph with $K$ edges. Define $\mathbf{\tilde{A}}, \bbA \in \{0,1\}^{(K+M)\times (K+M)}$ the zero padded adjacency matrices of their conflict graphs. The RGG can be seen as a perturbation of the DGG such that $\|\mathbf{\tilde{A}}-\bbA\|\leq \varepsilon$.\footnote{By embedding both graphs into a spatially-aligned vector space, the perturbation captures only the mismatched edges between the two topologies.} Consider a graph filter $\bbH$ that is integral Lipschitz with constant $C$ and whose coefficients are normalized. The difference between applying the filter to the two graphs with input signal $\bbx$ can be bounded as follows.
\begin{align}
    \|\bbH(\mathbf{\tilde{A}})\bbx-\bbH(\bbA)\bbx\|^2 \leq 4C\varepsilon \|\bbx\|.
\end{align}
\end{theorem}
\begin{proof}
    The extended adjacency matrices are symmetric and therefore allow their eigendecomposition to be written. As the difference between the matrices is $\|\mathbf{\tilde{A}}-\bbA\|\leq \varepsilon$, the difference between their eigenvalues is also bounded:
\begin{align}
    |\tilde{\lambda}_i-\lambda_i|\leq \varepsilon
\end{align}
Let $\delta>0$ small. For a given eigenvalue $\tilde{\lambda}_i$ of $\tilde{\bbA}$, we have eigenvalues ${\lambda}_{j}$ of $\mathbf{{A}}$ that are closer than $\delta$, such that $|\tilde{\lambda}_{i}-\lambda_j|\leq \delta$, and eigenvalues ${\lambda}_{j}$ of $\mathbf{{A}}$ that are further away than $\delta$, such that $|\tilde{\lambda}_{i}-\lambda_j|> \delta$. Let us use this to construct an eigenbasis via the eigenvalues of $\mathbf{{A}}$ as follows.
\begin{align}
    \bbV_i =& [{v}_j]_{j:|\tilde{\lambda}_{i}-\lambda_j|\leq \delta}\nonumber\\
    \bbV_i^C =& [{v}_j]_{j:|\tilde{\lambda}_{i}-\lambda_j|> \delta}
\end{align}
Let $\tilde{v_i}$ be an eigenvector of $\bbA$, we can write $\tilde{v_i} = \bbV_i\bbV_i^H\tilde{v_i}+\bbV_i^C(\bbV_i^{C})^H\tilde{v_i}$. We apply the graph filter to both graphs with input $\tilde{v_i}$ and compare the corresponding outputs. 
\begin{align}
    \bbH(\mathbf{\tilde{A}})\tilde{v_i}-\bbH(\bbA)\tilde{v_i} = (\bbH(\mathbf{\tilde{A}})-\bbH(\bbA))\tilde{v_i} = \Delta_i
\end{align}
Because the eigenvectors form a complete orthonormal basis, bounding the filter output difference for a given eigenvector $\tilde{v_i}$ is sufficient to bound the difference in the outputs for any arbitrary graph signal. Next, we consider the squared operator norm of $\Delta_i$. We replace the filter applied to $\tilde{\bbA}$, $\bbH(\tilde{\bbA})$ by the frequency response representation of the filter.
\begin{align}\label{eq:delta_i}
    \|\Delta_i\|^2 =& \|h(\tilde{\lambda}_{i})\tilde{v_i}-\bbH(\mathbf{{A}})\tilde{v_i}\|^2\nonumber\\
    =&\|h(\tilde{\lambda}_{i})\bbI-\bbH(\mathbf{{A}})\tilde{v_i}](\bbV_i\bbV_i^H\tilde{v_i}+\bbV_i^C(\bbV_i^{C})^H\tilde{v_i})\|^2\nonumber\\
    =&\|[h(\tilde{\lambda}_{i})\bbI-\bbH(\mathbf{{A}})\tilde{v_i}]\bbV_i\bbV_i^H\tilde{v_i}\|^2 \nonumber\\
    &+ \|[h(\tilde{\lambda}_{i})\bbI-\bbH(\mathbf{{A}})\tilde{v_i}]\bbV_i^C(\bbV_i^{C})^H\tilde{v_i}\|^2,
\end{align}
where the crossterms disappear because the vectors are orthogonal. Define the terms of Equation \eqref{eq:delta_i} $\Delta_i^1$ and $\Delta_i^2$, respectively. We have that $\Delta_i^1$ is associated to those eigenvalues of $\mathbf{{A}}$ that are closer than $\delta$ to the eigenvalue $\tilde{\lambda}_{i}$ of $\tilde{\bbA}$, while $\Delta_i^2$ considers those further away.

To bound $\Delta_i^1$ we expand $\bbH(\mathbf{{A}})$ via the eigendecomposition of $\mathbf{{A}}$.
\begin{align}
\Delta_i^1 &= \bbH(\mathbf{{A}})\bbV_i\bbV_i^H \tilde{v_i} - h(\tilde{\lambda_i})\bbV_i\bbV_i^H \tilde{v_i} \nonumber\\
&= \bbV h({\Lambda}) \bbV^H \bbV_i\bbV_i^H \tilde{v_i} - \bbV_i h(\tilde{\lambda_i})I \bbV_i^H \tilde{v_i} \nonumber\\
&= \bbV_i \left( h({\Lambda}_i) - h(\tilde{\lambda_i})\bbI \right) \bbV_i^H \tilde{v_i}
\end{align}
Let $h(\cdot)$ be an integral Lipschitz filter with constant $C$, we can bound $\Delta_i^1$ as follows.
\begin{align}
    \|\Delta_i^1\|^2 \leq C^2\delta^2,
\end{align}
where we consider the eigenvectors are orthonormal.

To bound $\Delta_i^2$, we apply the triangle inequality to the operator norm of the filter difference. Noting the filter coefficients are normalized such that $\|\bbH(\mathbf{{A}})\| = \max_j |h({\lambda}_j)| \leq 1$, we get:
\begin{align}\label{eq:firsttermdelta2}
    \|\bbH(\mathbf{{A}})-h(\tilde{\lambda_i})\bbI\|^2\leq& (\|\bbH(\mathbf{{A}})\|+\|h(\tilde{\lambda_i})\bbI\|)^2\nonumber\\
    \leq&(1 + 1)^2 = 4.
\end{align}
We must also bound the projection of $\tilde{v_i}$ onto the complement subspace, $\|(\bbV_i^C)^H\tilde{v_i}\|^2$. Using the fact that $(\mathbf{{A}} - \tilde{\bbA})\tilde{v_i} = (\mathbf{{A}} - \tilde{\lambda_i} \bbI)\tilde{v_i}$, we left-multiply by $(\bbV_i^C)^H$ to obtain:
\begin{align}
    (\bbV_i^C)^H(\mathbf{{A}}-\tilde{\bbA})\tilde{v_i} =& ({\Lambda}_i^C - \tilde{\lambda_i} \bbI)(\bbV_i^C)^H\tilde{v_i}
\end{align}

Taking the norm of both sides and applying the Davis-Kahan theorem, we bound the vector projection. Since the eigenvalues in ${\Lambda}_i^C$ are strictly further than $\delta$ from $\tilde{\lambda_i}$, we have:
\begin{align}\label{eq:secondtermdelta2}
    \|(\bbV_i^C)^H\tilde{v_i}\| \leq \frac{\|(\mathbf{{A}}-\tilde{\bbA})\tilde{v_i}\|}{\min_j |\tilde{\lambda_i} - {\lambda}_j^C|} \leq \frac{\varepsilon}{\delta}.
\end{align}
Putting Equations \eqref{eq:firsttermdelta2} and \eqref{eq:secondtermdelta2} together, we obtain a bound for $\Delta_i^2$:
\begin{align}
    \|\Delta_i^2\|^2 \leq& \|\bbH(\mathbf{{A}})-h(\tilde{\lambda_i})\bbI\|^2 \|(\bbV_i^C)^H\tilde{v_i}\|^2 \nonumber\\
    \leq& 4\frac{\varepsilon^2}{\delta^2}
\end{align}

Finally, substituting the bounds for both terms yields the final bound for the squared error:
\begin{align}\label{eq:prefinalbound}
    \|\Delta_i\|^2 =& \|\bbH(\mathbf{{A}})\tilde{v_i} - h(\tilde{\lambda_i})\tilde{v_i}\|^2\nonumber\\
    \leq& C^2\delta^2 + 4\frac{\varepsilon^2}{\delta^2}.
\end{align}
Let $\bbx\in\mathbb{R}^{(K+M)\times(K+M)}$ be the input to the graph filter, we can write $\bbx$ in terms of the eigenvectors $\tilde{v_i}$:
\begin{align}
    \bbx = \sum_{i=1}^n (\tilde{v_i}^H\bbx \tilde{v_i}) = \sum_{i=1}^n (\mathbf{\tdx}_i \tilde{v_i}).
\end{align}
To finish the proof, we bound the squared norm of $\Delta_i\bbx$ by observing the eigenvectors are orthogonal and therefore the cross products are zero.
\begin{align}
    \|\Delta_i\bbx\|_2^2 =& \|\Delta_i\sum_{i=1}^n (\mathbf{\tdx}_i \tilde{v_i})\|_2^2\nonumber \\
    =& \sum_{i=1}^n \|\Delta_i\mathbf{\tdx}_i \tilde{v_i}\|_2^2\nonumber \\
    =& \sum_{i=1}^n \mathbf{\tdx}_i^2\|\Delta_i \tilde{v_i}\|_2^2,
\end{align}
where we find that the first term is $\|\bbx\|_2^2$ and the second term is what we bounded in Equation \eqref{eq:prefinalbound}. Finally, if we define $\delta^2=\frac{2\varepsilon}{C}$, we get the final bound:
\begin{align}
    \|\bbH(\mathbf{{A}})\bbx-\bbH(\tilde{\bbA})\bbx\|^2 \leq 4C\varepsilon \|\bbx\|.
\end{align}
\end{proof}

Having shown the transferability of graph filters, we proceed to extend the result to Graph Neural Networks in Theorem \ref{th:rggtodgggnn}.
\begin{theorem}\label{th:rggtodgggnn}
    Let $\tdG$ be a DGG with $m_1$ edges and $\bbG$ be a RGG with $m_2$ edges. Define $\bbA, \tilde{\bbA} \in \{0,1\}^{(m_1+m_2)\times (m_1+m_2)}$ the zero padded adjacency matrices of their conflict graphs, such that $\|\bbA-\tilde{\bbA}\|\leq \varepsilon$. Let $\bbPhi(\mathbf{x},\mathbf{A};\mathbf{H})$ be a $L$-layer GNN with a graph filter $\bbH$ that is integral Lipschitz with constant $C$ and a normalized Lipschitz nonlinearity $\gamma$. The GNN can be transferred between $\tilde{\bbA}$ and $\bbA$ with minimal performance loss:
\begin{align}
    \|\bbPhi(\bbx, \tilde{\bbA}; \bbH)-\bbPhi(\bbx, \bbA; \bbH)\| \leq& 2L\sqrt{\varepsilon C}\|\bbx\|.
\end{align}
\end{theorem}
\begin{proof}
    
We compare the outputs of the GNN applied to both graphs via $\bbA$ and $\mathbf{\tilde{A}}$ for layer $l$, considering the pointwise nonlinearity $\sigma(\cdot)$ is normalized Lipschitz.
\begin{align}
    \|\bbPhi_l(\bbx, \mathbf{\tilde{A}}; \bbH)-\bbPhi_{l}(\bbx, \bbA; \bbH)\| =& \|\sigma(\bbH_l(\mathbf{\tilde{A}})\bbPhi_{l-1}(\bbx, \mathbf{\tilde{A}}; \bbH))\nonumber\\
    \quad &-\sigma(\bbH_l(\bbA)\bbPhi_{l-1}(\bbx, \bbA; \bbH))\| \nonumber\\
    \leq& \|\bbH_l(\mathbf{\tilde{A}})\bbPhi_{l-1}(\bbx, \mathbf{\tilde{A}}; \bbH)\nonumber \\
    \quad &-\bbH_l(\bbA)\bbPhi_{l-1}(\bbx, \bbA; \bbH)\|.
\end{align}
Add and subtract the cross term $\bbH_l(\mathbf{\tilde{A}})\bbPhi_{l-1}(\bbx, \bbA; \bbH)$. By applying the triangle inequality, we establish a recursion:
\begin{align}
    \|\bbPhi_l(\bbx, \mathbf{\tilde{A}}; \bbH)-\bbPhi_{l}(\bbx, \bbA; \bbH)\| \leq& \|\bbH_l(\mathbf{\tilde{A}})\bbPhi_{l-1}(\bbx, \mathbf{\tilde{A}}; \bbH)\nonumber\\
    \quad&-\bbH_l(\mathbf{\tilde{A}})\bbPhi_{l-1}(\bbx, \bbA; \bbH)\|\nonumber\\
    &+\|\bbH_l(\mathbf{\tilde{A}})\bbPhi_{l-1}(\bbx, \bbA; \bbH)\nonumber\\
    \quad&-\bbH_l(\bbA)\bbPhi_{l-1}(\bbx, \bbA; \bbH)\|\nonumber \\
    \leq& \|\bbH_l(\mathbf{\tilde{A}})\|\|\bbPhi_{l-1}(\bbx, \mathbf{\tilde{A}}; \bbH)\nonumber\\
    \quad&- \bbPhi_{l-1}(\bbx, \bbA; \bbH)\|\nonumber \\
    &+\|\bbH_l(\mathbf{\tilde{A}})-\bbH_l(\bbA)\|\nonumber\\
    \quad\quad&\|\bbPhi_{l-1}(\bbx, \bbA; \bbH)\|.
\end{align}
We assume that the filter coefficients are normalized, such that $\|\bbH_l(\mathbf{\tilde{A}})\|\leq 1$. Furthermore, the combination of non-amplifying filters and the normalized Lipschitz property of the activation function ensures that $\|\bbPhi_{l-1}(\bbx, \bbA; \bbH)\|\leq\|\bbx\|$. Incorporating the transferability bound of the single graph filter, we obtain:
\begin{align}
    \|\bbPhi_l(\bbx, \mathbf{\tilde{A}}; &\bbH)-\bbPhi_{l}(\bbx, \bbA; \bbH)\| \leq \nonumber\\
    &\|\bbPhi_{l-1}(\bbx, \mathbf{\tilde{A}}; \bbH)- \bbPhi_{l-1}(\bbx, \bbA; \bbH)\| + 2\sqrt{\varepsilon C}\|\bbx\|.
\end{align}
By unrolling the recursion over the layers, we arrive at the final bound for the outputs of a GNN with $L$ layers:
\begin{align}
    \|\bbPhi_L(\bbx, \mathbf{\tilde{A}}; \bbH)-\bbPhi_L(\bbx, \bbA; \bbH)\| \leq& 2L\sqrt{\varepsilon C}\|\bbx\|.
\end{align}
\end{proof}

\subsection{Proof of Theorem 2}
\label{app:the2}
\begin{theorem}\label{th:dggtondgg}
Let $\bbPhi(\bbx, {\tilde{\bbA}}_n; \bbH)$ be the parameterized policy that achieves a performance loss $ \ccalL({\tilde{\bbA}}_n,\bbH) $ when applied on a grid graph with size $n$ and achieves a loss of $ \ccalL({\tilde{\bbA}}_m,\bbH) $ when applied on another grid graph with size $m$. Suppose $n > m$, the difference of these two losses can be bounded as 
\begin{align}
    \ccalL({\tilde{\bbA}}_n,\bbH) \leq \ccalL({\tilde{\bbA}}_m,\bbH) + C\mathbb{E}[\bbx^2]+ 2\sqrt{\ccalL({\tilde{\bbA}}_m,\bbH)C\mathbb{E}[\bbx^2]},
\end{align}
where $C=\frac{H_K^2}{m}(2\sqrt{m}K+K^2)$ and $H_K=\sum_{k=0}^{K-1}|h_k|\|\tilde{\bbA}_m\|_2^k$.
\end{theorem}
\begin{proof}
Given we consider a 2D space, we can write the performance $\ccalL({\tilde{\bbA}}_m,\bbH)$ as the sum of the expected difference for each entry of the window of size $m$.
\begin{align}\label{eq:lb1}
    \ccalL({\tilde{\bbA}}_m,\bbH)=\frac{1}{m^2}\mathbb{E}\Big[\sum_{i,j=0}^{m-1} |\bbPhi(\bbx, \tilde{\bbA}_m; \bbH)(i, j)-\bby(i,j)|^2\Big].
\end{align}
We are interested in comparing $\ccalL({\tilde{\bbA}}_m,\bbH)$ as presented in \eqref{eq:lb1} to $\ccalL({\tilde{\bbA}}_n,\bbH)$. Let us define $\epsilon(i,j)=\bbPhi(\bbx, \tilde{\bbA}_n; \bbH)(i, j)-\bby(i,j)$ to simplify the notation. 
\begin{align}\label{eq:lb2}
    \ccalL({\tilde{\bbA}}_n,\bbH)=&\frac{1}{n^2}\mathbb{E}\Big[\sum_{i,j=0}^{n-1} |\bbPhi(\bbx, \tilde{\bbA}_n; \bbH)(i, j)-\bby(i,j)|^2\Big]\nonumber\\
    \leq&\frac{1}{(Nm)^2}\mathbb{E}\Big[\sum_{i,j=0}^{Nm-1} |\epsilon(i,j)|^2\Big],
\end{align}
where we defined $N=\lceil\frac{n}{m}\rceil$. We recenter the sums, denoting $\ccalT=\{im-\frac{(N-1)m}{2} \mid i\in\mathbb{Z}, 1\leq i<N\}$. This is equivalent to defining small squares in the 2D space in which we partition the sums, where each point has coordinates $\tau=(\tau_1, \tau_2)$:
\begin{align}
    \ccalL({\tilde{\bbA}}_n,\bbH)\leq&\frac{1}{(Nm)^2}\mathbb{E}\Big[\sum_{\tau\in\ccalT} \Big[\sum_{i,j=0}^{m-1} |\epsilon(i-\tau_1,j-\tau_2)|^2\Big]\Big].
\end{align}
The sum over the partitions $\ccalT$ does not affect the expectation and can therefore be taken outside. Furthermore, because the input and output are assumed to be jointly stationary, we have $\mathbb{E}[|\epsilon(i-\tau_1,j-\tau_2)|^2]=\mathbb{E}[|\epsilon(i,j)|^2]$. 
\begin{align}\label{eq:afterstationarity}
    \ccalL({\tilde{\bbA}}_n,\bbH)\leq&\frac{1}{(Nm)^2}\sum_{\tau\in\ccalT} \Big[\mathbb{E}\Big[\sum_{i,j=0}^{m-1} |\epsilon(i,j)|^2\Big]\Big].
\end{align}
In Equation \eqref{eq:afterstationarity}, the dependence on $\tau$ inside the expectation has been removed. This allows the computation of the sum over $\ccalT$ independently from the other terms. As $|\ccalT|=N^2$, we can simplify the terms as follows:
\begin{align}
    \ccalL({\tilde{\bbA}}_n,\bbH)\leq&\frac{N^2}{(Nm)^2}\mathbb{E}\Big[\sum_{i,j=0}^{m-1} |\epsilon(i,j)|^2\Big]\nonumber\\
    \leq&\frac{1}{m^2}\mathbb{E}\Big[\sum_{i,j=0}^{m-1} |\epsilon(i,j)|^2\Big].
\end{align}
Let $\sqcap_{m}$ be a masking operator acting on the larger space $\mathbb{R}^n$. For any spatial signal, $\sqcap_{m}$ pointwise multiplies the signal by the indicator function $\mathbf{1}((i,j)\in m\times m)$, zeroing out all entries outside the $m \times m$ window while preserving the dimensionality $n$. Under this unified dimension, we have $\bbx_m = \sqcap_{m}\bbx$, and  $\tilde{\bbA}_m = \sqcap_{m}\tilde{\bbA}_n\sqcap_{m}$. 

We now return to the original notation, dropping $\epsilon(i,j)$ and defining the full-scale output $\bbx_{n, L} = \bbPhi(\bbx, \tilde{\bbA}_n; \bbH)$ and the smaller-scale output $\bbx_{m, L} = \bbPhi(\sqcap_{m}\bbx, \tilde{\bbA}_m; \bbH)$. Because both are now naturally represented in $\mathbb{R}^n$, we can directly apply the masking operator to their difference.

\begin{align}\label{threeterms}
    \ccalL({\tilde{\bbA}}_n,\bbH)\leq&\frac{1}{m^2}\mathbb{E}\big[\|\sqcap_{m}(\bbx_{n,L}-\bby)\|^2\big]\nonumber\\
    \leq&\frac{1}{m^2}\mathbb{E}\big[\|\sqcap_{m}(\bbx_{n,L}-\bbx_{m,L})+\sqcap_{m}(\bbx_{m,L}-\bby)\|^2\big]\nonumber\\
    \leq&\frac{1}{m^2}\mathbb{E}\big[\|\sqcap_{m}(\bbx_{m,L}-\bby)\|^2\big]\nonumber\\
    &+\frac{1}{m^2}\mathbb{E}\big[\|\sqcap_{m}(\bbx_{n,L}-\bbx_{m,L})\|^2\big]\nonumber\\
    &+\frac{2}{m^2}\mathbb{E}\big[\|\sqcap_{m}(\bbx_{m,L}-\bby)\|\|\sqcap_{m}(\bbx_{n,L}-\bbx_{m,L})\|\big].
\end{align}
The first term in Equation \eqref{threeterms} is, by definition, $\ccalL({\tilde{\bbA}}_m,\bbH)$. Because $\bbx_{m,L}$ strictly operates within the $m$ region, applying the mask to it leaves it unchanged (i.e., $\sqcap_{m}\bbx_{m,L} = \bbx_{m,L}$). The second term can be bounded as follows:

\begin{align}
    \|\sqcap_{m}(\bbx_{n,L}-\bbx_{m,L})\|=& \Big\|\sqcap_{m}\Big(\sum_{k=0}^{K-1}h_k\tilde{\bbA}_n^k\bbx\Big) \nonumber\\
    &- \sum_{k=0}^{K-1}h_k(\sqcap_{m}\tilde{\bbA}_n\sqcap_{m})^k(\sqcap_{m}\bbx)\Big\|\nonumber\\
    \leq&\sum_{k=0}^{K-1}|h_k|\|\sqcap_{m}(\tilde{\bbA}_n^k\bbx) \nonumber \\
    \quad &- (\sqcap_{m}\tilde{\bbA}_n\sqcap_{m})^k(\sqcap_{m}\bbx)\|.
\end{align}
By the construction of the conflict graph, edges only exist between nodes that are at most adjacent or diagonal to each other. Therefore, a single application of the adjacency matrix spreads information by at most 1 hop. The error is bounded by the spectral norm of the adjacency matrix and the signal isolated at this boundary expansion:

\begin{align}
    \|\sqcap_{m}(\bbx_{n,L}-\bbx_{m,L})\|\leq&\sum_{k=0}^{K-1}|h_k|\|\tilde{\bbA}_n\|_2^k\|\sqcap_{m+k}(\bbx-\sqcap_{m}\bbx)\|\nonumber\\
    \leq&\sum_{k=0}^{K-1}|h_k|\|\tilde{\bbA}_n\|_2^k\|\sqcap_{m+K-1}(\bbx-\sqcap_{m}\bbx)\|.
\end{align}
Let $H_K=\sum_{k=0}^{K-1}|h_k|\|\tilde{\bbA}_n\|_2^k$. 
\begin{align}\label{eq:secondbound}
    \|\sqcap_{m}(\bbx_{n,L}-\bbx_{m,L})\|\leq& H_K\|\sqcap_{m+K-1}(\bbx-\sqcap_{m}\bbx)\|.
\end{align}
As $\bbx$ is stationary, the expected squared norm of the boundary term in Equation \eqref{eq:secondbound} can be evaluated as the variance of $\bbx$ over the volume difference $(m+K-1)^2-m^2$.

\begin{align}
    \mathbb{E}\big[\|\sqcap_{m}(\bbx_{n,L}-\bbx_{m,L})\|^2\big]\leq& H_K^2 \big((m+K-1)^2-m^2\big) \mathbb{E}[\bbx^2]\nonumber\\
    \leq& H_K^2(2mK+K^2)\mathbb{E}[\bbx^2].
\end{align}
This bounds the second term in Equation \eqref{threeterms}. Observe the third cross-term can also be bounded via Cauchy-Schwarz and the second term bound. Defining $C_1=H_K^2(2mK+K^2)$, we arrive at the final bound:

\begin{align}
    \ccalL({\tilde{\bbA}}_n,\bbH) \leq \ccalL({\tilde{\bbA}}_m,\bbH) + C_1\mathbb{E}[\bbx^2]+ 2\sqrt{\ccalL({\tilde{\bbA}}_m,\bbH)C_1\mathbb{E}[\bbx^2]}.
\end{align}    
\end{proof}

\subsection{Proof of Theorem 3}
\label{app:the3}
\begin{theorem}\label{the:rggtonrgg}
    Under assumptions analogous to Theorem \ref{the:rgg-gg-gnntransf}, consider an L-layer GNN $\bbPhi(\mathbf{x},\mathbf{A};\mathbf{H})$ trained to minimize the loss $\ccalL(\bbA_m,\bbH)$ for RGGs of size $m$, such that $\ccalL(\bbA_m,\bbH)\leq\zeta$ and suppose the deviation of RGG to its corresponding DGG is bounded by $\epsilon$. The GNN can be transferred to graphs of size $n$ with minimal performance deterioration:
\begin{align}
&\nonumber | \ccalL({ {\bbA}}_n,\bbH)  - \ccalL({ {\bbA}}_m,\bbH) |= \\ 
    & \mathcal{O}\Bigg(\sqrt{\zeta}\left(\sqrt{\epsilon}\|\mathbf{x}_{n}\| + \sqrt{\epsilon}\|\mathbf{x}_{m}\|\right) + \epsilon\|\mathbf{x}_{n}\|^2  + \epsilon\|\mathbf{x}_{m}\|^2 \Bigg) 
\end{align}
\end{theorem}
\begin{proof}
    Let us decompose the loss difference as follows.
    \begin{align}
        |\ccalL({ {\bbA}}_n,\bbH)  - \ccalL({ {\bbA}}_m,\bbH)| \leq& |\ccalL({ {\bbA}}_n,\bbH)  - \ccalL({ \tilde{\bbA}}_n,\bbH)|\nonumber\\
        &+|\ccalL({ \tilde{\bbA}}_m,\bbH)  - \ccalL({ \tilde{\bbA}}_n,\bbH)|\nonumber \\
        &+|\ccalL({ \tilde{\bbA}}_m,\bbH)  - \ccalL({ {\bbA}}_m,\bbH)|
    \end{align}

Observe that the first and third terms relate the performance for a RGG and a DGG that are sufficiently close, and therefore they can be bounded using Theorem \ref{the:rgg-gg-gnntransf}. The second term relates two grid graphs of different sizes and can be bounded via Theorem \ref{th:dggtondgg}. This concludes the proof.
\end{proof}



\end{document}